\documentclass[11pt,a4paper]{article}
\usepackage[hyperref]{acl2020}
\usepackage{times}
\usepackage{latexsym}
\usepackage[textsize=tiny]{todonotes}
\usepackage{enumitem}
\usepackage{tabularx}
\usepackage{multirow}
\usepackage{soul}

\usepackage{booktabs}
\usepackage{placeins}

\usepackage{microtype}

\aclfinalcopy %
\def\aclpaperid{951} %

\title{TACRED Revisited: A Thorough Evaluation of the TACRED Relation Extraction Task}

\author{Christoph Alt ~~~~~ Aleksandra Gabryszak ~~~~~
Leonhard Hennig \\
\mbox{}\\
German Research Center for Artificial Intelligence (DFKI)\\
Speech and Language Technology Lab \\
\texttt{\{christoph.alt, aleksandra.gabryszak, leonhard.hennig\}@dfki.de}}

\date{}

\begin{document}
\maketitle

\setlength{\tabcolsep}{5pt}

\newcommand{\tableRelabelAgreement}{
    \begin{table}[ht!]
    \centering
        \begin{tabular}{@{}lcc|cc@{}}
        \toprule
        & \multicolumn{2}{c}{Dev} & \multicolumn{2}{c}{Test} \\
        IAA &   H1,H2 & H,C & H1,H2 & H,C \\
        \midrule
        \emph{Challenging} & 0.78 & 0.43 & 0.85  &  0.44\\ 
        \emph{Control} &  0.87 & 0.95 & 0.94 & 0.96 \\
        \midrule
        \emph{All} &  0.80 & 0.53 & 0.87 & 0.55 \\ 
        \bottomrule
        \end{tabular}
    \caption{Inter-Annotator Kappa-agreement for the relation validation task on TACRED \emph{Dev} and \emph{Test} splits (H1,H2 = human re-annotators, H = revised labels, C = original TACRED crowd-generated labels).}
        \label{tab:relabelAgreement}
    \end{table}
}

\newcommand{\tableRelabelStatistics}{
\begin{table*}[t!]
    \centering
    \begin{tabular}{@{}lrr|rr@{}}
    \toprule
      & \multicolumn{2}{c}{Dev} & \multicolumn{2}{c}{Test} \\
      & \emph{Challenging} & \emph{Control} & \emph{Challenging} & \emph{Control} \\
    \midrule
    \# Examples (\# positive) & 3,088 (1,987) & 567 (547) & 1,923 (1,333) & 427 (407) \\
    
    \# Revised (\# positive) &  1,610 \phantom{x}(976) & 46 \phantom{x}(46) & 960 \phantom{x}(630) & 38 \phantom{x}(38) 
    
    \\\midrule %
    
    \# Revised (\% positive) & \textbf{52.1} (\textbf{49.1}) & \textbf{8.1} (\textbf{8.4})&  \textbf{49.9} (\textbf{47.3}) & \textbf{8.9} (\textbf{9.3}) \\ %
    \bottomrule
    \end{tabular}
    \caption{Re-annotation statistics for TACRED \emph{Dev} and \emph{Test} splits.}
    \label{tab:relabelStatistics}
    \end{table*}
}

\newcommand{\tableFAutomated}{
\begin{table}[t]
    \centering
\begin{tabular}{@{}lccc@{}}
\toprule
{} & Original & Revised & Weighted \\
Model        &          &          &          \\
\midrule
CNN, masked  & 59.5 & 66.5 & 34.8 \\
TRE          & 67.4 & 75.3 & 48.8 \\
SpanBERT     & 70.8 & 78.0 & 61.9 \\
KnowBERT     & 71.5 & 79.3 & 58.7 \\
\bottomrule
\end{tabular}
\caption{Test set F1 score on TACRED, our revised version, and weighted by difficulty (on revised). The weight per instance is determined by the number of incorrect predictions in our set of 49 RE models. The result suggests that SpanBERT better generalizes to more challenging examples, e.g. complex sentential context.}
    \label{tab:f1score}
    \end{table}
}

\newcommand{\insertreannotationresultstable}{
    \begin{table}[ht!]\centering
        \begin{tabular}{@{}lcc@{}}
            \toprule
            Error & Dev & Test \\
            \midrule
            Annotation Errors & x & x \\
            Model Errors & x & x \\
            Other Errors & x & x \\
            \bottomrule
        \end{tabular}
    \caption{x}
    \label{tab:reannotation_results}
    \end{table}
}

\newcommand{\insertmisclfcategoryexampletablenew}{
    \begin{table*}[ht!]
    \centering
    \small
        \begin{tabular}{@{}lp{0.635\textwidth}p{0.1\textwidth}r@{}}
            \toprule
            \multicolumn{1}{c}{\textbf{Error Type}} & 
            \multicolumn{1}{c}{\textbf{Examples}} & 
            \multicolumn{1}{c}{\textbf{Prediction}} &
            \multicolumn{1}{c}{\textbf{Freq.}}
            \\\midrule
            \textbf{Arguments} & & \\
            Span & This is a tragic day for the \underline{Australian [Defence Force]}$_{head:org}$ ([ADF]$_{tail:org}$)  &\emph{org:alt.\_nam} & 12 \\
            Entity Type & [Christopher Bollyn]$_{head:per}$ is an [\underline{independent}]$_{tail:religion}$ journalist \newline The company, which [Baldino]$_{head:org}$ founded in [1987]$_{tail:date}$ sells a variety of drugs  &  \emph{per:religion} \newline \emph{org:founded}& 31 \\  
            
             \midrule
            \textbf{Context} & & \\
            Inverted Args & [\underline{Ruben van Assouw}]$_{head:per}$, who had been on safari with his 40-year-old father [\underline{Patrick}]$_{tail:per}$ , mother Trudy , 41 , and brother Enzo , 11 . & \emph{per:children} & 25 \\
            Wrong Args &  Authorities said they ordered the detention of \underline{Bruno 's wife} , [Dayana Rodrigues]$_{tail:per}$ , who was found with [Samudio]$_{head:per}$'s baby . &  \emph{per:spouse} & 109 \\
            Ling. Distractor & In May , [he]$_{head:per}$ secured \$ 96,972 in \underline{working} capital from [GE Healthcare Financial Services]$_{tail:org}$ . & \emph{per:employ.\_of} & 35 \\
            Factuality  & 
            [Ramon]$_{head:per}$ said he \underline{hoped to one day become} an [astronaut]$_{head:title}$ \newline \underline{Neither he nor} [Aquash]$_{head:per}$ were [American]$_{tail:nationality}$ citizens . &  \emph{per:title} \newline  \emph{per:origin}  & 11 \\
            
            Relation Def. & [Zhang Yinjun]$_{tail:per}$ , \underline{spokesperson} with one of China 's largest charity organization , the [China Charity Federation]$_{head:org}$ & \emph{org:top\_mem.} & 96 \\ %
            Context Ignored & [Bibi]$_{head:per}$ , a mother of [five]$_{tail:number}$, was sentenced this month to death . &  \emph{per:age} & 52 \\ %
            No Relation & [He]$_{head:per}$ turned a gun on himself committing [suicide]$_{tail:cause of death }$ . & \emph{no\_relation} & 646\\
            \midrule
            {} & {} & \textbf{Total} & \textbf{1017} \\ %
           \bottomrule
        \end{tabular}
    \caption{Misclassification types along with sentence examples, relevant false predictions, and error frequency. The problematic sentence parts are underlined (examples may be abbreviated due to space constraints).}
    \label{tab:misclf_cat_examples}
    \end{table*}
}

\newcommand{\inserttacredstatstable}{
    \begin{table}[t!]\centering
        \begin{tabular}{@{}lcc@{}}
            \toprule
            Split & \# Examples & \# Neg. examples\\
            \midrule
            Train & 68,124 & 55,112 \\
            Dev & 22,631 & 17,195 \\
            Test & 15,509 & 12,184 \\
            \bottomrule
        \end{tabular}
    \caption{TACRED statistics per split. About 79.5\% of the examples are labeled as \emph{no\_relation}.}
    \label{tab:tacred_dataset_stats}
    \end{table}
}

\newcommand{\inserttacredmodelsetresults}{
\begin{table*}[h]
\small
\centering
    \begin{tabular}{@{}lccc|ccc@{}}
    \toprule
    {} & \multicolumn{3}{c}{Original} & \multicolumn{3}{c}{Revised} \\
    {} &        P &    R &   F1 &       P &    R &   F1 \\
    Model                                &          &      &      &         &      &      \\
    \midrule
    BoE                                  & 50.0 & 32.6 & 39.4 & 51.8 & 35.9 & 42.4 \\
    CNN                                  & 72.3 & 45.5 & 55.9 & 79.8 & 53.5 & 64.1 \\
    CNN, masked                          & 67.2 & 53.5 & 59.5 & 72.5 & 61.4 & 66.5 \\
    CNN w/ POS/NER                       & 72.2 & 54.7 & 62.2 & 79.7 & 64.3 & 71.2 \\
    CNN + ELMo                           & 73.8 & 48.8 & 58.8 & 82.1 & 57.9 & 67.9 \\
    CNN + ELMo, masked                   & 72.3 & 53.8 & 61.7 & 79.8 & 63.2 & 70.5 \\
    CNN + ELMo, masked w/ POS/NER        & 69.2 & 59.0 & 63.7 & 76.0 & 69.1 & 72.4 \\
    CNN + BERT uncased                   & 71.9 & 51.1 & 59.7 & 79.5 & 60.2 & 68.5 \\
    CNN + BERT uncased, masked           & 69.0 & 62.0 & 65.3 & 74.9 & 71.7 & 73.2 \\
    CNN + BERT cased                     & 69.7 & 54.3 & 61.0 & 77.6 & 64.3 & 70.4 \\
    CNN + BERT cased, masked             & 71.8 & 61.1 & 66.1 & 78.1 & 70.8 & 74.3 \\
    LSTM                                 & 59.3 & 47.5 & 52.7 & 65.9 & 56.2 & 60.6 \\
    LSTM, masked                         & 63.4 & 51.7 & 57.0 & 68.7 & 59.7 & 63.9 \\
    LSTM, masked w/ POS/NER              & 65.4 & 56.8 & 60.8 & 71.2 & 66.0 & 68.5 \\
    LSTM + ELMo                          & 61.5 & 61.3 & 61.4 & 68.1 & 72.2 & 70.1 \\
    LSTM + ELMo, masked                  & 63.9 & 64.9 & 64.4 & 69.3 & 75.0 & 72.1 \\
    LSTM + ELMo, masked w/ POS/NER       & 61.7 & 67.8 & 64.6 & 66.1 & 77.3 & 71.2 \\
    LSTM + BERT uncased                  & 64.7 & 60.2 & 62.4 & 71.6 & 71.0 & 71.3 \\
    LSTM + BERT uncased, masked          & 65.3 & 64.8 & 65.1 & 70.4 & 74.3 & 72.3 \\
    LSTM + BERT cased                    & 66.2 & 59.8 & 62.8 & 73.5 & 70.8 & 72.1 \\
    LSTM + BERT cased, masked            & 68.9 & 61.9 & 65.2 & 75.0 & 71.8 & 73.4 \\
    Bi-LSTM                              & 53.3 & 57.4 & 55.3 & 58.6 & 67.2 & 62.6 \\
    Bi-LSTM, masked                      & 62.5 & 63.4 & 62.9 & 67.7 & 73.1 & 70.3 \\
    Bi-LSTM + ELMo                       & 65.0 & 58.7 & 61.7 & 72.6 & 69.8 & 71.1 \\
    Bi-LSTM + ELMo, masked               & 63.3 & 64.8 & 64.1 & 68.9 & 75.2 & 71.9 \\
    Bi-LSTM + ELMo w/ POS/NER            & 64.8 & 57.9 & 61.2 & 72.1 & 68.6 & 70.3 \\
    Bi-LSTM + ELMo, masked w/ POS/NER    & 63.0 & 65.9 & 64.4 & 67.5 & 75.2 & 71.2 \\
    Bi-LSTM + BERT uncased               & 65.3 & 59.9 & 62.5 & 71.8 & 70.2 & 71.0 \\
    Bi-LSTM + BERT uncased, masked       & 64.9 & 66.0 & 65.4 & 69.6 & 75.3 & 72.4 \\
    Bi-LSTM + BERT cased                 & 65.2 & 61.2 & 63.1 & 72.1 & 72.1 & 72.1 \\
    Bi-LSTM + BERT cased, masked         & 68.3 & 64.0 & 66.1 & 74.1 & 73.9 & 74.0 \\
    GCN                                  & 65.6 & 50.5 & 57.1 & 72.4 & 59.3 & 65.2 \\
    GCN, masked                          & 68.2 & 58.0 & 62.7 & 74.3 & 67.4 & 70.7 \\
    GCN, masked w/ POS/NER               & 68.6 & 60.2 & 64.2 & 74.2 & 69.3 & 71.7 \\
    GCN + ELMo                           & 66.5 & 57.6 & 61.7 & 73.4 & 67.7 & 70.4 \\
    GCN + ELMo, masked                   & 68.5 & 61.3 & 64.7 & 74.5 & 71.0 & 72.7 \\
    GCN + ELMo, masked w/ POS/NER        & 67.9 & 64.8 & 66.3 & 73.3 & 74.4 & 73.9 \\
    GCN + BERT uncased                   & 66.3 & 58.8 & 62.4 & 73.1 & 69.1 & 71.0 \\
    GCN + BERT uncased, masked           & 68.7 & 64.0 & 66.3 & 74.8 & 74.1 & 74.5 \\
    GCN + BERT cased                     & 66.5 & 56.4 & 61.0 & 74.4 & 67.1 & 70.5 \\
    GCN + BERT cased, masked             & 67.2 & 64.6 & 65.9 & 72.9 & 74.7 & 73.8 \\
    S-Att.                               & 56.9 & 58.3 & 57.6 & 62.2 & 67.8 & 64.9 \\
    S-Att., masked                       & 65.0 & 66.8 & 65.9 & 69.3 & 75.8 & 72.4 \\
    S-Att. + ELMo                        & 64.4 & 65.0 & 64.7 & 71.5 & 76.8 & 74.1 \\
    S-Att. + ELMo, masked                & 64.0 & 69.4 & 66.6 & 68.9 & 79.6 & 73.8 \\
    S-Att. + BERT uncased                & 60.6 & 67.6 & 63.9 & 66.3 & 78.7 & 72.0 \\
    S-Att. + BERT uncased, masked        & 64.0 & 69.7 & 66.7 & 68.9 & 80.0 & 74.0 \\
    S-Att. + BERT cased                  & 63.5 & 64.1 & 63.8 & 70.4 & 75.7 & 73.0 \\
    S-Att. + BERT cased, masked          & 69.2 & 64.7 & 66.9 & 75.1 & 74.8 & 75.0 \\
    \midrule
    Average                              & 65.6 & 59.5 & 62.1 & 71.8 & 69.2 & 70.1 \\
    \bottomrule
    \end{tabular}
\caption{Test set performance on TACRED and the revised version for all 49 models we used to select the most challenging instances. We use the same entity masking strategy as \citet{zhang-etal-2017-position}, replacing each entity in the original sentence with a special \textless NER\textgreater -\{SUBJ, OBJ\} token where \textless NER\textgreater\ is the corresponding NER tag. For models w/ POS/NER we concatenate part-of-speech and named entity tag embeddings to each input token embedding.}
\label{tab:test_performance_tacred_models}
\end{table*}
}

\begin{abstract}
TACRED~\cite{zhang-etal-2017-position} is one of the largest, most widely used crowdsourced datasets in Relation Extraction (RE). But, even with recent advances in unsupervised pre-training and knowledge enhanced neural RE, models still show a high error rate. In this paper, we investigate the questions: Have we reached a performance ceiling or is there still room for improvement? And how do crowd annotations, dataset, and models contribute to this error rate?
To answer these questions, we first validate the most challenging 5K examples in the development and test sets using trained annotators. We find that label errors account for 8\% absolute F1 test error, and that more than 50\% of the examples need to be relabeled. On the relabeled test set the average F1 score of a large baseline model set improves from 62.1 to 70.1.
After validation, we analyze misclassifications on the challenging instances, categorize them into linguistically motivated error groups, and verify the resulting error hypotheses on three state-of-the-art RE models. We show that two groups of ambiguous relations are responsible for most of the remaining errors and that models may adopt shallow heuristics on the dataset when entities are not masked.

\end{abstract}

\section{Introduction}

\begin{figure}[t!]
\centering
    \includegraphics[width=\linewidth]{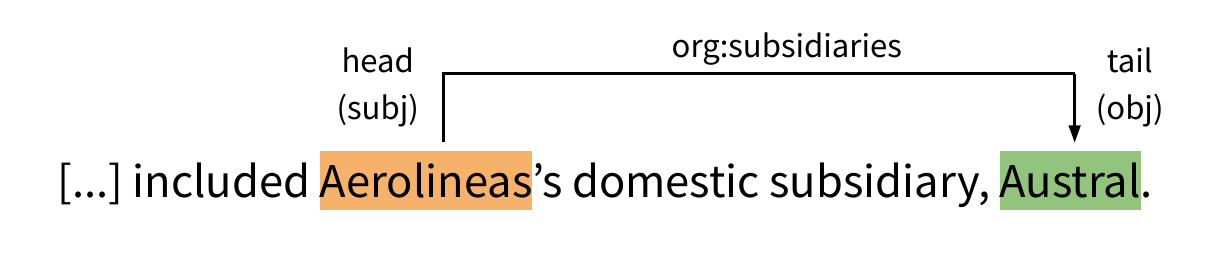}
    \caption{Relation example from TACRED. The sentence contains the relation \emph{org:subsidiaries} between the head and tail organization entities `Aerolineas' and `Austral'.}
    \label{fig:re_example}
\end{figure}

Relation Extraction (RE) is the task of extracting relationships between concepts and entities from text, where relations correspond to semantic categories such as \textit{per:spouse}, \emph{org:founded\_by} or \textit{org:subsidiaries} (Figure \ref{fig:re_example}). This makes RE a key part of many information extraction systems, and its performance determines the quality of extracted facts for knowledge base population \cite{tac_kbp_2010}, or the quality of answers in question answering systems \cite{re_for_qa}. Standard benchmarks such as SemEval 2010 Task 8 \cite{hendrickx-etal-2010-semeval} and the more recent TACRED \cite{zhang-etal-2017-position} are essential to evaluate new RE methods and their limitations, and to establish open challenges.

TACRED is one of the largest and most widely used RE datasets. It contains more than 106k examples annotated by crowd workers. The methods best performing on the dataset use some form of pre-training to improve RE performance: fine-tuning pre-trained language representations \cite{alt-etal-2019-improving,shi-etal-2019-simple,joshi-etal-2019-spanbert} or integrating external knowledge during pre-training, e.g.\ via joint language modelling and linking on entity-linked text \cite{zhang-etal-2019-ernie,peters-etal-2019-knowledge,baldini-soares-etal-2019-matching}; with the last two methods achieving a state-of-the-art performance of 71.5 F1. While this performance is impressive, the error rate of almost 30\% is still high. The question we ask in this work is: Is there still room for improvement, and can we identify the underlying factors that contribute to this error rate? We analyse this question from two separate viewpoints: (1) to what extent does the quality of crowd based annotations contribute to the error rate, and (2) what can be attributed to dataset and models?
Answers to these questions can provide insights for improving crowdsourced annotation in RE, and suggest directions for future research.

To answer the first question, we propose the following approach: We first rank examples in the development and test sets according to the misclassifications of 49 RE models and select the top 5k instances for evaluation by our linguists. This procedure limits the manual effort to only the most challenging examples.
We find that a large fraction of the examples are mislabeled by the crowd. Our first contribution is therefore a extensively relabeled TACRED development and test set.

To answer the second question, we carry out two analyses: (1) we conduct a manual explorative analysis of model misclassifications on the most challenging test instances and categorize them into several linguistically motivated error categories; (2) we formulate these categories into testable hypotheses, which we can automatically validate on the full test set by adversarial rewriting -- removing the suspected cause of error and observing the change in model prediction~\cite{wu-etal-2019-errudite}. We find that two groups of ambiguous relations are responsible for most of the remaining errors. The dataset also contains clues that are exploited by models without entity masking, e.g.\ to correctly classify relations even with limited access to the sentential context.

We limit our analysis to TACRED, but want to point out that our approach is applicable to other RE datasets as well. We make the code of our analyses publicly available.\footnote{\url{https://github.com/DFKI-NLP/tacrev}} In summary, our main contributions in this paper are:
\begin{itemize}
    \item We validate the 5k most challenging examples in the TACRED development and test sets, and provide a revised dataset\footnote{Due to licensing restrictions, we can not publish the dataset but instead provide a patch to the original TACRED.} that will improve the accuracy and reliability of future RE method evaluations.
    \item We evaluate the most challenging, incorrectly predicted examples of the revised test set, and develop a set of 9 categories for common RE errors, that will also aid evaluation on other datasets.
    \item We verify our error hypotheses on three state-of-the-art RE models and show that two groups of ambiguous relations are responsible for most of the remaining errors and that models exploit cues in the dataset when entities are unmasked.
\end{itemize}

\section{The TACRED Dataset}
\label{sec:tacred-v1}

The \textit{TAC} \textit{R}elation \textit{E}xtraction \textit{D}ataset\footnote{\url{https://catalog.ldc.upenn.edu/LDC2018T24}}, introduced by \citet{zhang-etal-2017-position}, is a fully supervised dataset of sentence-level binary relation mentions. It consists of 106k sentences with entity mention pairs collected from the TAC KBP\footnote{\url{https://tac.nist.gov/2017/KBP/index.html}} evaluations 2009--2014, with the years 2009 to 2012 used for training, 2013 for development, and 2014 for testing. 
Each sentence is labeled with one of 41 person- and organization-oriented relation types, e.g.\ \emph{per:title}, \emph{org:founded}, or the label \emph{no\_relation} for negative instances. Table~\ref{tab:tacred_dataset_stats} summarizes key statistics of the dataset. %

\inserttacredstatstable

All relation labels were obtained by crowdsourcing, using Amazon Mechanical Turk. Crowd workers were shown the example text, with head (subject) and tail (object) mentions highlighted, and asked to select among a set of relation label suggestions, or to assign \emph{no\_relation}. Label suggestions were limited to relations compatible with the head and tail types.\footnote{See the supplemental material provided by~\citet{zhang-etal-2017-position} for details of the dataset creation and annotation process.}

The data quality is estimated as relatively high by~\citet{zhang-etal-2017-position}, based on a manual verification of 300 randomly sampled examples (93.3\% validated as correct). The inter-annotator kappa label agreement of crowd workers was moderate at $\kappa = 0.54$ for 761 randomly selected mention pairs.

\section{An Analysis of TACRED Label Errors}
\label{sec:relabel_augment}
In order to identify the impact of potentially noisy, crowd-generated labels on the observed model performance, we start with an analysis of  TACRED's label quality. We hypothesize that while comparatively untrained crowd workers may on average produce relatively good labels for easy relation mentions, e.g.\ those with obvious syntactic and/or lexical triggers, or unambiguous entity type signatures such as \emph{per:title}, they may frequently err on challenging examples, e.g.\ highly ambiguous ones or relation types whose scope is not clearly defined.

An analysis of the complete dataset using trained annotators would be prohibitively expensive. We therefore utilize a principled approach to selecting examples for manual analysis (Section~\ref{subsec:identify_challenging}). Based on the TAC-KBP annotation guidelines, we then validate these examples (Section~\ref{subsec:relation_relabeling}), creating new \emph{Dev} and \emph{Test} splits where incorrect annotations made by crowd workers are revised (Section~\ref{subsec:tacred_2}).

\subsection{Data Selection}
\label{subsec:identify_challenging}
Since we are interested in identifying potentially incorrectly labeled examples, we implement a selection strategy which is based upon ordering examples by the difficulty of predicting them correctly.\footnote{A similar approach was used e.g.\ by \citet{barnes-etal-2019-sentiment}.} We use a set of 49 different RE models to obtain predictions on the development and test sets,
and rank each example according to the number of models predicting a different relation label than the ground truth.\footnote{See the supplemental material for details on the models, training procedure, hyperparameters, and task performance.} %
Intuitively, examples with large disagreement, between all models or between models and the ground truth, are either difficult, or incorrectly annotated. 

We select the following examples for validation: (a) \emph{Challenging} -- all examples that were misclassified by at least half of the models, and (b) \emph{Control} -- a control group of (up to) 20 random examples per relation type, including \emph{no\_relation}, from the set of examples classified correctly by at least 39 models. The two groups cover both presumably hard and easy examples, and allow us to contrast validation results based on example difficulty. In total we selected 2,350 (15.2\%) \emph{Test} examples and 3,655 (16.2\%) \emph{Dev} examples for validation. Of these, 1,740 (\emph{Test}) and 2,534 (\emph{Dev}) were assigned a positive label by crowd workers.

\subsection{Human Validation}
\label{subsec:relation_relabeling}
We validate the selected examples on the basis of the TAC KBP guidelines.\footnote{\url{https://tac.nist.gov/2014/KBP/ColdStart/guidelines/TAC_KBP_2014_Slot_Descriptions_V1.4.pdf}} We follow the approach of~\citet{zhang-etal-2017-position}, and present each example by showing the example's text with highlighted head and tail spans, and a set of relation label suggestions.  We differ from their setup by showing more label suggestions to make the label choice less restrictive: (a) the original, crowd-generated ground truth label, (b) the set of labels predicted by the models, (c) any other relation labels matching the head and tail entity types, and (d) \emph{no\_relation}. 
The suggested positive labels are presented in an alphabetical order and are followed by \emph{no\_relation}, with no indication of a label's origin. Annotators are asked to assign \emph{no\_relation} or up to two positive labels from this set. A second label was allowed only if the sentence expressed two relations, according to the guidelines, e.g.\ \emph{per:city\_of\_birth} and \emph{per:city\_of\_residence}.
Any disagreements are subsequently resolved by a third annotator, who is also allowed to consider the original ground truth label. All annotators are educated in general linguistics, have extensive prior experience in annotating data for information extraction tasks, and are trained in applying the task guidelines in a trial annotation of 500 sentences selected from the development set.

\subsection{The Revised TACRED Dev and Test Sets}
\label{subsec:tacred_2}
\tableRelabelStatistics

Table~\ref{tab:relabelStatistics} shows the results of the validation process. In total, the annotators revised 960 (49.9\%) of the \emph{Challenging Test} examples, and 1,610 (52.1\%) of the \emph{Challenging Dev} examples, a very large fraction of label changes for both dataset splits. Revision rates for originally positive examples are lower at 47.3\% (\emph{Test}) and 49.1\% (\emph{Dev}). Approximately 57\% of the negative examples were relabeled with a positive relation label (not shown). Two labels were assigned to only 3.1\% of the \emph{Test}, and 2.4\% of the \emph{Dev} examples. The multi-labeling mostly occurs with location relations, e.g.\ the phrase \emph{``[Gross]$_{head:per}$, a 60-year-old native of [Potomac]$_{tail:city}$''} is labeled with \emph{per:city\_of\_birth} 
and \emph{per:city\_of\_residence}, which is justified by the meaning of the word \emph{native}.

As expected, the revision rate in the \emph{Control} groups is much lower, at 8.9\% for \emph{Test} and 8.1\% for \emph{Dev}.
We can also see that the fraction of negative examples is approximately one-third in the \emph{Challenging} group, much lower than the dataset average of 79.5\%. This suggests that models have more difficulty predicting positive examples correctly.

\tableRelabelAgreement
The validation inter-annotator agreement is shown in Table~\ref{tab:relabelAgreement}. It is very high at $\kappa_{Test} = 0.87$ and $\kappa_{Dev}=0.80$, indicating a high annotation quality. For both \emph{Test} and \emph{Dev}, it is higher for the easier \emph{Control} groups than for the \emph{Challenging} groups. In contrast, the average agreement between our annotators and the crowdsourced labels is much lower at $\kappa_{Test} = 0.55, \kappa_{Dev} = 0.53$, and lowest for \emph{Challenging} examples (e.g., $\kappa_{Test} = 0.44$).

Frequently erroneous crowd labels are \emph{per:cities\_of\_residence}, \emph{org:alternate\_names}, and \emph{per:other\_family}. Typical errors include mislabeling an example as positive which does not express the relation, e.g.\ labeling 
\emph{``[Alan Gross]$_{head:per}$ was arrested at the [Havana]$_{tail:loc}$ airport.''}
as \emph{per:cities\_of\_residence}, or not assigning a positive relation label, e.g. \emph{per:other\_family} in 
\emph{``[Benjamin Chertoff]$_{head:per}$ is the Editor in Chief of Popular Mechanics magazine, as well as the cousin of the Director of Homeland Security, [Michael Chertoff]$_{tail:per}$''}.
Approximately 49\% of the time an example's label was changed to \emph{no\_relation} during validation, 36\% of the time from \emph{no\_relation} to a positive label, and the remaining 15\% it was changed to or extended with a different relation type.

To measure the impact of dataset quality on the performance of models, we evaluated all 49 models on the revised test split. The average model F1 score rises to 70.1\%, a major improvement of 8\% over the 62.1\% average F1 on the original test split, corresponding to a 21.1\% error reduction.

\paragraph{Discussion}
The large number of label corrections and the improved average model performance show that the quality of crowdsourced annotations is a major factor contributing to the overall error rate of models on TACRED. Even though our selection strategy was biased towards examples challenging for models, the large proportion of changed labels suggests that these examples were difficult to label for crowd workers as well. To put this number into perspective -- \citet{riedel-modeling-2010} showed that, for a distantly supervised dataset, about 31\% of the sentence-level labels were wrong, which is less than what we observe here for human-supervised data.\footnote{Riedel et al.'s estimate is an average over three relations with 100 randomly sampled examples each, for similar news text. Two of the relations they evaluated, \emph{nationality} and \emph{place\_of\_birth}, are also contained in TACRED, the third is \emph{contains} (location).}

The low quality of crowd-generated labels in the \emph{Challenging} group may be due to their complexity, or due to other reasons, such as lack of detailed annotation guidelines, lack of training, etc. It suggests that, at least for \emph{Dev} and \emph{Test} splits, crowdsourcing, even with crowd worker quality checks as used by~\citet{zhang-etal-2017-position}, may not be sufficient to produce high quality evaluation data. While models may be able to adequately utilize noisily labeled data for training, measuring model performance and comparing progress in the field may require an investment in carefully labeled evaluation datasets. This may mean, for example, that we need to employ well-trained annotators for labeling evaluation splits, or that we need to design better task definitions and task presentations setups as well as develop new quality control methods when using crowd-sourced annotations for complex NLP tasks like RE.

\section{An Analysis of Model Errors}
 After revising the dataset, we focus on the two open questions: which of the remaining errors can be attributed to the models, and what are potential reasons for misclassifications? To answer these, we first create an annotation task instructing the linguists to annotate model misclassifications with their potential causes (Section~\ref{subsec:misclf_annotation}). We then categorize and analyze the causes and formulate testable hypotheses that can be automatically verified (Section~\ref{subsec:hypotheses_and_rewrite}). For the automatic analysis, we implemented a baseline and three state-of-the-art models (Section~\ref{subsec:models}).

\subsection{Misclassification Annotation}
\label{subsec:misclf_annotation}
The goal of the annotation is to identify possible linguistic aspects that cause incorrect model predictions. We first conduct a manual exploratory analysis on the revised \textit{Control} and \textit{Challenging} test instances that are misclassified by the majority of the 49 models. Starting from single observations, we iteratively develop a system of categories based on the existence, or absence, of contextual and entity-specific features that might mislead the models (e.g. entity type errors or distracting phrases). Following the exploration, we define a final set of categories, develop guidelines for each, and instruct two annotators to assign an error category to each misclassified instance in the revised test subset. In cases where multiple categories are applicable the annotator selected the most relevant one. As in the validation step, any disagreements between the two annotators are resolved by a third expert. %

\subsection{Error Hypotheses Formulation and Adversarial Rewriting}
\label{subsec:hypotheses_and_rewrite}
In a next step, we extend the misclassification categories to testable hypotheses, or groups, that are verifiable on the whole dataset split. For example, if we suspect a model to be distracted by an entity in context of same type as one of the relation arguments, we formulate a group \textit{has\_distractor}. The group contains all instances, both correct and incorrect, that satisfy a certain condition, e.g.\ there exists at least one entity in the sentential context of same type as one of the arguments. The grouping ensures that we do not mistakenly prioritize groups that are actually well-handled on average. We follow the approach proposed by~\citet{wu-etal-2019-errudite}, and extend their Errudite framework\footnote{\url{https://github.com/uwdata/errudite}} to the relation extraction task. After formulating a hypothesis, we assess the error prevalence over the entire dataset split to validate whether the hypothesis holds, i.e. the group of instances shows an above average error rate. In a last step, we test the error hypothesis explicitly by adversarial rewriting of a group's examples, e.g.\ by replacing the distracting entities and observing the models' predictions on the rewritten examples. In our example, if the \textit{has\_distractor} hypothesis is correct, removing the entities in context should change the prediction of previously incorrect examples.

\subsection{Models}
\label{subsec:models}
We evaluate our error hypotheses on a baseline and three of the most recent state-of-the-art RE models. None of the models were part of the set of models used for selecting challenging instances (Section~\ref{subsec:identify_challenging}), so as not to bias the automatic evaluation. As the baseline we use a single layer CNN \cite{zeng-etal-2014-relation, nguyen-grishman-2015-relation} with max-pooling and 300-dimensional GloVe \cite{pennington-etal-2014-glove} embeddings as input. The state-of-the-art models use pre-trained language models (LM) fine-tuned to the RE task and include: \textbf{TRE} \cite{alt-etal-2019-improving}, which uses the unidirectional OpenAI Generative Pre-Trained Transformer (GPT) \cite{radford-etal-2018-improvinglu}; \textbf{SpanBERT} \cite{joshi-etal-2019-spanbert}, which employs a bidirectional LM similar to BERT~\cite{devlin-etal-2019-bert} but is pre-trained on span-level; and \textbf{KnowBERT}~\cite{peters-etal-2019-knowledge}, which is an extension to BERT that integrates external knowledge. In particular, we use KnowBERT-W+W, which is trained by joint entity linking and language modelling on Wikipedia and WordNet. %

\section{Model Error and Dataset Analysis}
\label{sec:automated-analysis}
In this section, we present our analysis results, providing an answer to the question: which of the remaining errors can be attributed to the models, and what are the potential reasons for these errors? We first discuss the findings of our manual misclassification analysis (Section~\ref{subsec:misclf_results}), followed by the results of the automatic analysis (Section~\ref{subsec:automated_analysis}).

\subsection{Model Error Categories}
\label{subsec:misclf_results}
\insertmisclfcategoryexampletablenew

Table~\ref{tab:misclf_cat_examples} summarizes the linguistic misclassification categories we developed. We distinguish between errors resulting from (1) relation argument errors, and (2) context misinterpretation.\footnote{The manual analysis focused on the sentence semantics, and left aspects such such as sentence length, distance between entities, etc.\ for the automatic analysis, which can handle the analysis of surface features more effectively.} 
The category \textit{relation argument errors} refers to misclassifications resulting from incorrectly assigned \textit{entity spans} or \textit{entity types} of arguments. We always labeled type annotation errors, but tolerated minor span annotation errors if they did not change the interpretation of the relation or the entity.

The category \textit{context misinterpretation} refers to cases where 
the sentential context of the arguments is misinterpreted by the model. We identify the following context problems:
(1)~\textit{Inverted arguments}: the prediction is inverse to the correct relation, i.e.\ the model's prediction would be correct if head and tail were swapped.
(2)~\textit{Wrong arguments}: the model incorrectly predicts a relation that holds between head or tail and an un-annotated entity mention in the context, therefore misinterpreting one annotated argument.
(3)~\textit{Linguistic distractor}: the example contains words or phrases related to the predicted relation, however they do not connect to any of the arguments in a way justifying the prediction.
(4)~\textit{Factuality}: the model ignores negation, speculation, future tense markers, etc.
(5)~\textit{Context ignored}: the example does not contain  sufficient linguistic evidence for the predicted relation except for the matching entity types. 
(6)~\textit{Relation definition}: the predicted relation could be inferred from the context using common sense or world knowledge, however the inference is prohibited by the guidelines (e.g.\ the spokesperson of an organization is not a top member/employee, or a work location is not a pointer to the employee's residence). 
(7)~\textit{No Relation}: the model incorrectly predicts \emph{no\_relation} even though there is sufficient linguistic evidence for the relation in the sentential context.

\paragraph{Discussion}

The  relation label predicted most frequently across the 49 models disagreed with the ground truth label of the re-annotated \emph{Challenging} and \emph{Control Test} groups in 1017 (43.3\%) of the cases. The inter-annotator agreement of error categories assigned to these examples is high at $\kappa_{Test} = 0.83$ ($\kappa_{Test} = 0.67$ if the category \emph{No Relation} is excluded). %

Argument errors accounted for only 43 (4.2\%) misclassifications, since the entities seem to be mostly correctly assigned in the dataset. In all entity type misclassification cases except one, the errors originate from false annotations in the dataset itself.

\begin{figure*}[t!]
    \centering
    \includegraphics[width=\linewidth,trim={5 5 5 5},clip]{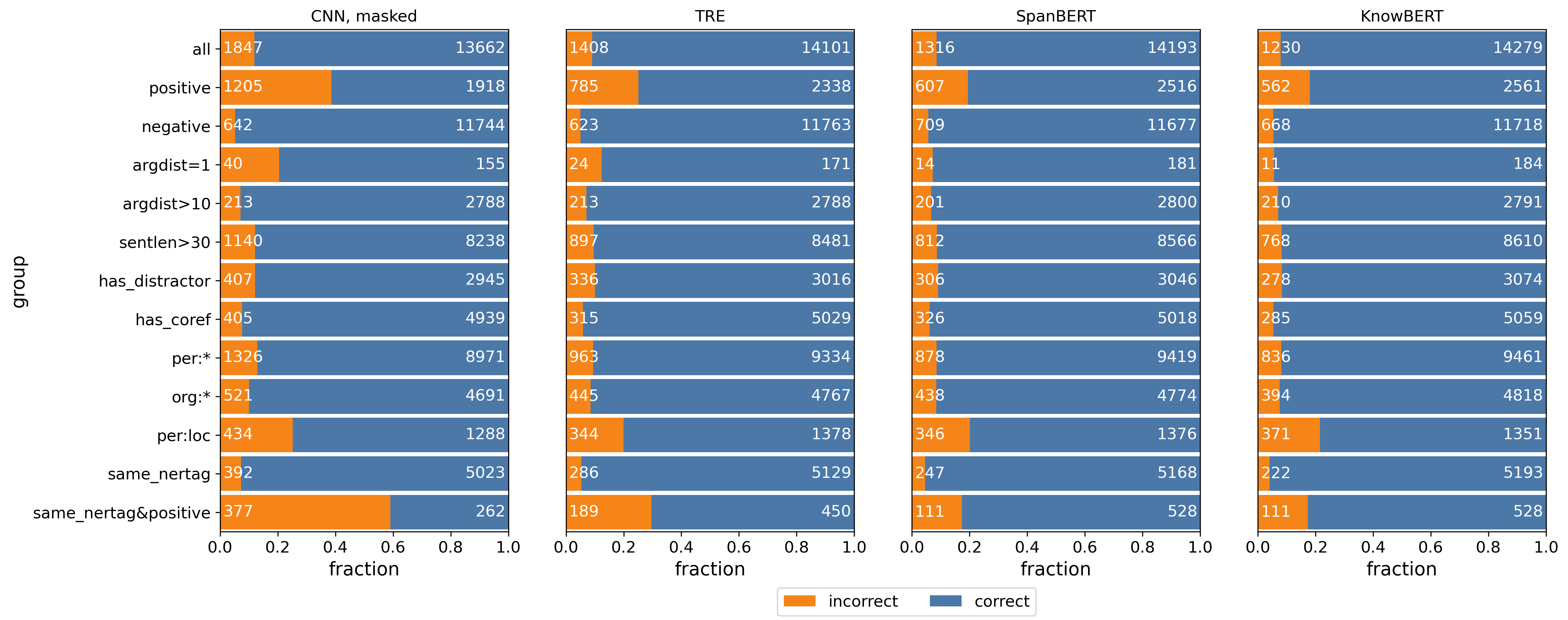}
    \caption{Error rates for different groups (example subsets) on the revised TACRED test set, for four different models. The bars show the number and fraction of correctly (blue) and incorrectly (orange) predicted examples in the given group. KnowBert, as the best-performing model, has the lowest error rates across most groups. Error rates for \emph{per:loc}, \emph{same\_nertag\&positive} are higher for all models than the model error rate on the complete test set (\emph{all}), highlighting examples for further error analysis and potential model improvements.}
    \label{fig:error_rates_per_group}
\end{figure*}

Context misinterpretation caused 974 (95.8\%) false predictions. %
\emph{No relation} is incorrectly assigned in 646 (63.6\%) of misclassified instances, even though the correct relation is often explicitly and unambiguously stated. %
In 134 (13.2\%) of the erroneous instances the misclassification resulted from  inverted or wrong argument assignment, i.e.\ the predicted relation is stated, however the arguments are inverted or the predicted relation involves an entity other than the annotated one.
In 96 (9.4\%) instances the error results from TAC KBP guidelines prohibiting specific inferences, affecting most often the classification of the relations \emph{per:cities\_of\_residence} and \emph{org:top\_member/employee}. %
Furthermore, in 52 (5.1\%) of the false predictions models seem to ignore the sentential context of the arguments, i.e.\ the predictions are inferred mainly from the entity types.
Sentences containing linguistic distractors accounted for 35 (3.4\%) incorrect predictions. 
Factuality recognition causes only 11 errors (1.1\%). However, we assume that this latter low error rate is due to TACRED data containing an insufficient number of sentences suitable for extensively testing a model's ability to consider the missing factuality of relations.

\subsection{Automatic Model Error Analysis}
\label{subsec:automated_analysis}
For the automatic analysis, we defined the following categories and error groups:
\begin{itemize}
    \item \emph{Surface structure} -- Groups for argument distance (\emph{argdist=1, argdist\textgreater10}) and sentence length (\emph{sentlen\textgreater30})
    \item \emph{Arguments} -- Head and tail mention NER type (\emph{same\_nertag, per:*, org:*, per:loc}), and pronominal head/tail (\emph{has\_coref})
    \item \emph{Context} -- Existence of distracting entities (\emph{has\_distractor})
    \item \emph{Ground Truth} -- Groups conditioned on the ground truth (\emph{positive, negative, same\_nertag\&positive})
\end{itemize}
Figure~\ref{fig:error_rates_per_group} shows the error rates of different groups on the revised TACRED test set. The plot shows error rates across four representative models. Each chart displays groups on the y-axis, and the fraction and number of correct (blue) vs. incorrect (orange) instances in the respective group on the x-axis. The average error rate of each model on the full test set is shown for reference in the top-most column titled \emph{all}. Groups with higher than average error rate may indicate characteristics of examples that make classification difficult. On the other hand, groups with lower than average error rate comprise examples that the given model performs especially well on.

\paragraph{What is the error rate for different groups?}
In Figure~\ref{fig:error_rates_per_group}, we can see that KnowBERT has the lowest error rate on the full test set (7.9\%), and the masked CNN model the highest (11.9\%). SpanBERT's and TRE's error rates are in between the two. Overall, all models exhibit a similar pattern of error rates across the groups, with KnowBERT performing best across the board, and the CNN model worst. 
We can see that model error rates e.g.\ for the groups \emph{has\_distractor}, \emph{argdist\textgreater10}, and \emph{has\_coref} do not diverge much from the corresponding overall model error rate. The presence of distracting entities in the context therefore does not seem to be detrimental to model performance. Similarly, examples with a large distance between the relation arguments, or examples where co-referential information is required, are generally predicted correctly. %
On the other hand, we can see that all models have above-average error rates for the group \emph{positive}, its subgroup \emph{same\_nertag\&positive}, and the group \emph{per:loc}. The above-average error rate for \emph{positive} may be explained by the fact that the dataset contains much fewer positive than negative training instances, and is hence biased towards predicting \emph{no\_relation}.
A detailed analysis shows that the groups \emph{per:loc} and \emph{same\_nertag\&positive} are the most ambiguous. \emph{per:loc} contains relations such as \emph{per:cities\_of\_residence}, \emph{per:countries\_of\_residence} and \emph{per:origin}, that may be expressed in a similar context but differ only in the fine-grained type of the tail argument (e.g. per:city vs. per:country). In contrast, \emph{same\_nertag} contains all person-person relations such as \emph{per:parents}, \emph{per:children} and \emph{per:other\_family}, as well as e.g.\ \emph{org:parent} and \emph{org:subsidiaries} that involve the same argument types (per:per vs. org:org) and may be only distinguishable from context.

\paragraph{How important is context?}
KnowBERT and SpanBERT show about the same error rate on the groups \emph{per:loc} and \emph{same\_nertag\&positive}. They differ, however, in which examples they predict correctly: For \emph{per:loc}, 78.6\% are predicted by both models, and  21.4\% are predicted by only one of the models. For \emph{same\_nertag\&positive}, 12.8\% of the examples are predicted by only of the models. The two models thus seem to identify complementary information. One difference between the models is that KnowBERT has access to entity information, while SpanBERT masks entity spans.

To test how much the two models balance context and argument information, we apply rewriting to alter the instances belonging to a group and observe the impact on performance. We use two strategies: (1) we remove all tokens outside the span between head and tail argument (\emph{outside}), and (2) we remove all tokens between the two arguments (\emph{between}). We find that SpanBERT's performance on \emph{per:loc} drops from 62.1 F1 to 57.7 (outside) and 43.3 (between), whereas KnowBERT's score decreases from 63.7 F1 to 60.9 and 50.1, respectively. On \emph{same\_nertag\&positive}, we observe a drop from 89.2 F1 to 58.2 (outside) and 47.7 (between) for SpanBERT. KnowBERT achieves a score of 89.4, which drops to 83.8 and 49.0. The larger drop in performance on \emph{same\_nertag\&positive} suggests that SpanBERT, which uses entity masking, focuses more on the context, whereas KnowBERT focuses on the entity content because the model has access to the arguments. Surprisingly, both models show similar performance on the full test set (Table~\ref{tab:f1score}). This suggests that combining both approaches may further improve RE performance.

\paragraph{Should instance difficulty be considered?}
Another question is whether the dataset contains instances that can be solved more easily than others, e.g.\ those with simple patterns or patterns frequently observed during training. We assume that these examples are also more likely to be correctly classified by our baseline set of 49 RE models.

To test this hypothesis, we change the evaluation setup and assign a weight to each instance based on the number of correct predictions. An example that is correctly classified by all 49 baseline models would receive a weight of zero -- and thus effectively be ignored -- whereas an instance misclassified by all models receives a weight of one.
In Table~\ref{tab:f1score}, we can see that SpanBERT has the highest score on the weighted test set (61.9 F1), a 16\% decrease compared to the unweighted revised test set. KnowBERT has the second highest score of 58.7, 3\% less than SpanBERT. The performance of TRE and CNN is much worse at 48.8 and 34.8 F1, respectively. The result suggests that SpanBERT's span-level pre-training and entity masking are beneficial for RE and allow the model to generalize better to challenging examples. Given this observation, we propose to consider an instance's difficulty during evaluation.

\tableFAutomated

\section{Related Work}

\textbf{Relation Extraction on TACRED}
Recent RE approaches include PA-LSTM~\cite{zhang-etal-2017-position} and GCN~\cite{zhang-etal-2018-graph}, with the former combining recurrence and attention, and the latter leveraging graph convolutional neural networks. Many current approaches use unsupervised or semi-supervised pre-training: fine-tuning of language representations pre-trained on token-level \cite{alt-etal-2019-improving,shi-etal-2019-simple} or span-level \cite{joshi-etal-2019-spanbert}, fine-tuning of knowledge enhanced word representations that are pre-trained on entity-linked text \cite{zhang-etal-2019-ernie, peters-etal-2019-knowledge}, and ``matching the blanks'' pre-training \cite{baldini-soares-etal-2019-matching}.

\textbf{Dataset Evaluation}
\citet{chen-etal-2016-thorough} and \citet{barnes-etal-2019-sentiment} also use model results to assess dataset difficulty for reading comprehension and sentiment analysis. Other work also explores bias in datasets and the adoption of shallow heuristics on biased datasets in natural language inference \cite{niven-kao-2019-probing} and argument reasoning comprehension \cite{mccoy-etal-2019-right}.

\textbf{Analyzing trained Models}
Explanation methods include occlusion or gradient-based methods, measuring the relevance of input features to the output \cite{zintgraf2017visualizing, harbecke-etal-2018-learning}, and probing tasks \cite{conneau-etal-2018-cram,kim-etal-2019-probing} that probe the presence of specific features e.g.\ in intermediate layers. More similar to our approach is rewriting of instances \cite{jia-liang-2017-adversarial,ribeiro-etal-2018-semantically} but instead of evaluating model robustness we use rewriting to test explicit error hypotheses, similar to \citet{wu-etal-2019-errudite}.

\section{Conclusion and Future Work}
In this paper, we conducted a thorough evaluation of the TACRED RE task. We validated the 5k most challenging examples in development and test set and showed that labeling is a major error source, accounting for 8\% absolute F1 error on the test set. This clearly highlights the need for careful evaluation of development and test splits when creating datasets via crowdsourcing. To improve the evaluation accuracy and reliability of future RE methods, we provide a revised, extensively relabeled TACRED.
In addition, we categorized model misclassifications into 9 common RE error categories and observed that models are often unable to predict a relation, even if it is expressed explicitly. Models also frequently do not recognize argument roles correctly, or ignore the sentential context.
In an automated evaluation we verified our error hypotheses on the whole test split and showed that two groups of ambiguous relations are responsible for most of the remaining errors. We also showed that models adopt heuristics when entities are unmasked and proposed that evaluation metrics should consider an instance's difficulty.

\section*{Acknowledgments}
We would like to thank all reviewers for their thoughtful comments and encouraging feedback, and Matthew Peters for providing KnowBERT predictions on TACRED. We would also like to thank Elif Kara, Ulli Strohriegel and Tatjana Zeen for the annotation of the dataset. This work has been supported by the German Federal Ministry of Education and Research as part of the projects DEEPLEE (01IW17001) and BBDC2 (01IS18025E), and by the German Federal Ministry for Economic Affairs and Energy as part of the project PLASS (01MD19003E).

\bibliography{main}
\bibliographystyle{acl_natbib}

\appendix
\section{Appendix}

\subsection{Hyperparameters}

\paragraph{CNN}
For training we use the hyperparameters of \citet{zhang-etal-2017-position}. We employ \textit{Adagrad} as an optimizer, with an initial learning rate of 0.1 and run training for 50 epochs. Starting from the 15th epoch, we gradually decrease the learning rate by a factor of 0.9. For the CNN we use 500 filters of sizes [2, 3, 4, 5] and apply $l_2$ regularization with a coefficient of $10^{-3}$ to all filter weights. We use tanh as activation and apply dropout on the encoder output with a probability of 0.5. We use the same hyperparameters for variants with ELMo. For variants with BERT, we use an initial learning rate of 0.01 and decrease the learning rate by a factor of 0.9 every time the validation F1 score is plateauing. Also we use 200 filters of sizes [2, 3, 4, 5].

\paragraph{LSTM/Bi-LSTM}
For training we use the hyperparameters of \citet{zhang-etal-2017-position}. We employ \textit{Adagrad} with an initial learning rate of 0.01, train for 30 epochs and gradually decrease the learning rate by a factor of 0.9, starting from the 15th epoch. We use word dropout of 0.04 and recurrent dropout of 0.5. The BiLSTM consists of two layers of hidden dimension 500 for each direction. For training with ELMo and BERT we decrease the learning rate by a factor of 0.9 every time the validation F1 score is plateauing.

\paragraph{GCN}
We reuse the hyperparameters of \citet{zhang-etal-2018-graph}. We employ \textit{SGD} as optimizer with an initial learning rate of 0.3, which is reduced by a factor of 0.9 every time the validation F1 score plateaus. We use dropout of 0.5 between all but the last GCN layer, word dropout of 0.04, and embedding and encoder dropout of 0.5. Similar to the authors we use path-centric pruning with K=1. We use two 200-dimensional GCN layers and similar two 200-dimensional feedforward layers with ReLU activation.

\paragraph{Self-Attention}
After hyperparameter tuning we found 8 layers of multi-headed self-attention to perform best. Each layer uses 8 attention heads with attention dropout of 0.1, keys and values are projected to 256 dimensions before computing the similarity and aggregated in a feedforward layer with 512 dimensions. For training we use \textit{Adam} optimizer with an initial learning rate of $10^{-4}$, which is reduced by a factor of 0.9 every time the validation F1 score plateaus. In addition we use word dropout of 0.04, embedding dropout of 0.5, and encoder dropout of 0.5.

\subsection{Relation Extraction Performance}
Table~\ref{tab:test_performance_tacred_models} show the relation extraction performances for the models on TACRED and our revised version. Models with `w/synt/sem' use named entity and part-of-speech embeddings in addition to the input word embeddings.

\inserttacredmodelsetresults

\end{document}